\documentclass[runningheads]{llncs}
\usepackage{graphicx}
\usepackage{subfig} 
\usepackage{adjustbox}
\usepackage{amsfonts}
\usepackage{bm}

\usepackage[subtle,margins=normal,leading=normal]{savetrees} 
\begin{document}
%
%
\title{Multi-task Federated Learning for \\Heterogeneous Pancreas Segmentation}
%
%
\author{Chen Shen \inst{1} \and
Pochuan Wang \inst{2} \and
Holger R. Roth \inst{3} \and
Dong Yang \inst{3} \and \\
Daguang Xu \inst{3} \and
Masahiro Oda \inst{1} \and
Weichung Wang \inst{2} \and
Chiou-Shann Fuh \inst{2} \and \\
Po-Ting Chen \inst{4} \and
Kao-Lang Liu \inst{4} \and
Wei-Chih Liao \inst{4} \and
Kensaku Mori \inst{1}
}
%
\authorrunning{C. Shen et al.}
\titlerunning{Multi-task Federated Learning for Heterogeneous Pancreas Segmentation}
%
\institute{Nagoya University, Japan \and
National Taiwan University, Taiwan \and
NVIDIA Corporation, United States \and
National Taiwan University Hospital, Taiwan
}
%
\maketitle              
\begin{abstract}
Federated learning (FL) for medical image segmentation becomes more challenging in multi-task settings where clients might have different categories of labels represented in their data. For example, one client might have patient data with ``healthy'' pancreases only while datasets from other clients may contain cases with pancreatic tumors. The vanilla federated averaging algorithm makes it possible to obtain more generalizable deep learning-based segmentation models representing the training data from multiple institutions without centralizing datasets. However, it might be sub-optimal for the aforementioned multi-task scenarios. In this paper, we investigate heterogeneous optimization methods that show improvements for the automated segmentation of pancreas and pancreatic tumors in abdominal CT images with FL settings.
\end{abstract}
\section{Introduction}
Fully automated segmentation of the pancreas and pancreatic tumors from CT volumes is still challenging due to the low contrast and significant variations across subjects caused by different scanning protocols and patient populations. A great deal of progress has been made to improve the pancreas segmentation performance in the last decade with the rapid development of convolutional neural network (CNN) based approaches to this medical image segmentation task \cite{roth2015deeporgan,zhou2017fixed,oktay2018attention,man2019deep}. Still, highly accurate and generalizable pancreas and corresponding tumor segmentation models are encouraged as a prerequisite for computer-aided diagnosis (CAD) systems. 

One limitation for pancreas segmentation using deep learning-based approaches is the lack of annotated training data. Distinct from natural images, collecting extensive training data from various resources may lead to multiple technical, legal, and privacy issues in healthcare applications. Federated learning (FL) is an innovative technique that enables to learn a deep learning-based model among distributed devices, i.e. clients, collaboratively without having to centralize the training data in one location \cite{McMahan2017}.

FL techniques currently attract increasing attention in the medical image analysis field because the acquisition of annotated medical images from the real world is challenging and costly. A growing number of studies have been made to handle these difficulties by using the FL strategy. Furthermore, FL shows great effectiveness on segmentation tasks in abdominal organs \cite{wang2020Automated,czeizler2020using}, brain tumors \cite{sheller2019Multiinstitutional,li2019Privacy} and COVID-19 image analysis \cite{dou2021Federated,flores2021Federated,xia2021AutoFedAvg,yang2021federated} both in simulation and real-world FL applications.

One question remaining is on how to integrate best models trained on heterogeneous tasks among clients is difficult to determine. Most commonly, the Federated Averaging (FedAvg) is used to aggregate the model from each client and to update the global model on the server using a weighted sum where the weights are typically derived from the local training dataset sizes and kept constant during the training \cite{McMahan2017}. FedProx was introduced to handle data heterogeneity in FL by adding a regularization loss on the client that penalizes divergence from the current global model \cite{li2018federated}. Techniques proposed for multi-task learning can be an alternative to tackle the heterogeneous statistical problems \cite{smith2017federated}. Dynamic task prioritization (DTP) specifies a prioritization for each task in multi-task learning based on the task-specific metrics\cite{guo2018dynamic}. Dynamic weight averaging (DWA) investigates a weight for each task calculated through the change of loss \cite{Liu_2019_CVPR}.

This work employs three public annotated datasets for pancreas segmentation to model three heterogeneous clients during FL. One dataset consists of pancreas and tumors, and the other two are consist of healthy pancreas cases. Our main contributions are as follows: 1) introduce the dynamic task prioritization for FL optimization; 2) investigate dynamic weight averaging aggregation method to re-weight the model from each client. 3) compare the effect of our improvements with FedAvg and FedProx on the pancreas and tumor segmentation task.

\section{Methods}
\label{sec:method}
\begin{figure}[tb]
    \centering
    \includegraphics[width=0.78\textwidth]{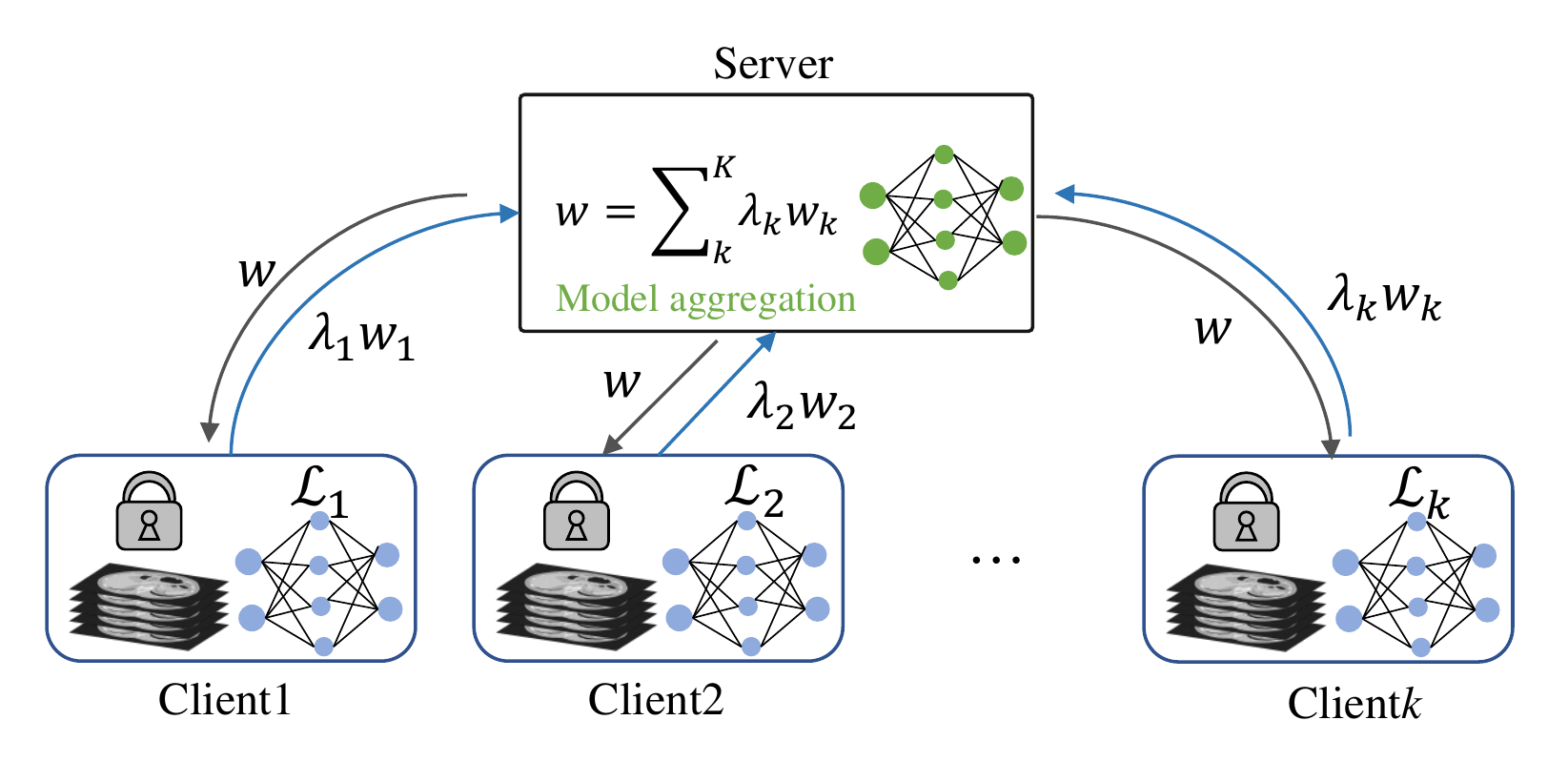}
    \caption{An illustration of federated learning in medical imaging. The server only receives model updates and the training data stays on the client sites privately.}
    \label{fig:fl}
\end{figure}

A standard FL system consists of a server and several clients. In a new FL round, each client receives the global model from the server and fine-tunes it on their local dataset. Then the client only shares a weight update with the server after the local training. The server is designed to receive the model updates from the specified minimum of clients and aggregates the updates based on the aggregation weight of each client. It then updates the global model with the aggregated updates and distributes the updated global model for the next round of FL training. An illustration of the FL system is shown in Fig. \ref{fig:fl}. 
The standard FL tries to minimize 
\begin{equation}
    \mathcal{L} = min\sum_{k}^K \eta_k \mathcal{L}_k,
\end{equation}
where the $k$-th client tries to optimize the local loss function $\mathcal{L}_k$. The total number of clients is $K$ and the proportion that each client contributes to the global model update is $\eta_k\geq0$, where $\sum_{k}^K\eta_k=1$. In this work, our aim is to investigate a multi-task FL optimization method for heterogeneous pancreas segmentation where clients exhibit different types of images and labels in their data.

In this section, we first revisit the FedAvg \cite{McMahan2017} and FedProx \cite{li2018federated} methods which are widely used in FL tasks. Then, we adapt two optimization methods from the multi-task learning literature to the FL setting: \textit{dynamic task prioritization} and \textit{dynamic weight averaging}.

\subsection{FedAvg}
\label{sec:fedavg}
In standard FedAvg, to reduce the heavy communication cost and to handle dropping clients, only a subset $K$ clients are used, instead of total $N$ clients to update the global model in each round. Here, we have $K<<N$. The weight $\eta_k$ of client $k\in K$ is a constant number which can be calculated by
\begin{equation}
\eta_k =\frac{n_k}{n},
\end{equation}
where $n_k$ is the number of local training data in client $k$. The total number of training data in all clients can be derived from ${n}=\sum_{k}^Kn_k$. In FedAvg, the client with larger local training data contributes more to the updated global model.

\subsection{FedProx}
\label{sec:fedprox}
FedProx is an improved federated optimization algorithm for learning from distributed heterogeneous datasets~\cite{li2018federated}. The FedProx algorithm is an extension of the standard FedAvg scheme.
The FedProx algorithm adopted the aggregation scheme and added another learning constraint for each client, namely a regularization The regularization term can help the local client model to stay close to server model. The local client try to minimize
\begin{equation}
    \hat{\mathcal{L}}_k=\mathcal{L}_{k}+\frac{\mu}{2}\left\| \bm{w}_k- \bm{w}\right\|^{2},
\end{equation}
where $\hat{\mathcal{L}}_k$ specifies the learning target of client $k$, and $\bm{w}_k$ stands for the local model parameters.The $\bm{w}$ is the model parameter from the FL global model, and $\left\|\cdot\right\|^2$ indicates the L2 normalization. As mentioned in section~\ref{sec:method}, $\mathcal{L}_{k}$ is the local loss function.

This learning constraint ensures the consistency of gradients from different clients. The more consistent gradient can prevent model divergence of client models and improve the convergence of the global model.
\subsection{Dynamic Task Prioritization}
\label{sec:dtp}
Dynamic task prioritization (DTP) for multi-task learning adjusts the weights between different tasks by estimating the key performance index (KPI) $\kappa$. $\kappa$ is a monotone increasing function ranged from 0 to 1; the larger value of $\kappa$ stands for better performance of the specific task. DTP concentrates on challenging tasks by increasing corresponding weights and lowering the weights of easier tasks.
We generalized the DTP for federated learning by considering each client as a different task. In this work, we define the KPI of client $k$ as
\begin{equation}
    \kappa_{k,i}=d_{k,i}^r,
\end{equation}
\begin{equation}
    W_{k,i} = -(1 - \bar{\kappa_{k,i}})^{\gamma}\log{\bar{\kappa_{k,i}}}.
\end{equation}
To stabilize the weights between each batch, an exponential average $\bar{\kappa_{k,i}}$ is used:

\begin{equation}
    \bar{\kappa_{k,i}} = \left( 1 - \alpha \right)\kappa_{k,i} + \alpha \bar{\kappa_{k,i-1}}.
\end{equation}
Here, $\alpha$ is a number between $0$ and $1$ and $\gamma$ is a tunable hyperparameter.

\subsection{Dynamic Weight Averaging}
In dynamic weight averaging (DWA), we try to optimize the FL procedure by focusing on server model aggregation instead of applying a constraint on loss function. This method is inspired by optimization approaches from classical multi-task learning tasks~\cite{Liu_2019_CVPR}. In FL, finding a suitable balance to aggregate the model updates from heterogeneous clients is challenging. 
However, to specify the proper weight requires a large number of experiments and priority knowledge. In DWA, we investigate a method that defines the client weights on each round automatically, The server learns to weigh each client based on the variation of loss values from the previous round. The weight of client $k$ in round $r$ can be define as
\begin{equation}
    \lambda_{k,r} = \frac{\xi \exp(\rho_{k,r-1}/T)}{\sum_{i=1}^K \exp(\rho_{i,r-1}/T)},
\end{equation}
where $\rho_{k,r-1} \in (0, +\infty)$ represents the dynamic proportion of the loss value $\mathcal{L}$ changes in client $k$ from the round before previous round $r-2$ to the previous round $r-1$, which can be defined as $\rho_{k,r-1} = {\mathcal{L}_{k,r-1}}/{\mathcal{L}_{k,r-2}}$. To control the effeteness of dynamic proportion, $T$ is defined as similar in the MTL in \cite{Liu_2019_CVPR}. When $T \to +\infty$, the weight of each client is tend to be equally $\rho_k \to 1$. We introduce $\xi\in\mathbb{N}$ to adjust the impact of weights in DWA.
Different from the way to calculate loss value $ \mathcal{L}_{k,r}$ in \cite{Liu_2019_CVPR}, we average the local loss value of each iteration in one round, which can be defined as
\begin{equation}
    \mathcal{L}_{k,r} = \frac{1}{J}\sum_{j=1}^{J} \mathcal{L}_{k, j},
\end{equation}
where $j$ is the local iteration number within the total $J$ iterations. The average operation will make the loss value of each round more stable. For the first round (when $r=1$), we initialize the ${\mathcal{L}_{k,r-1}}=1, {\mathcal{L}_{k,r-2}}=1$ so that we can have the $\rho_{k,1} = 1$ after the first round.

\section{Experiments and Results}
\subsection{Datasets}

The experiment is conducted with one federated server for model aggregation and three clients for training. The server does not own any validation data and only aggregates the client's model parameters. Each client owns a dataset from a different source. The first dataset is the Pancreas-CT from The National Institutes of Health Clinical Center (TCIA)\footnote{\url{https://wiki.cancerimagingarchive.net/display/Public/Pancreas-CT}}~\cite{roth2015deeporgan}. This dataset contains 82 abdominal contrast-enhanced CT scans with manual segmentation labels for the healthy pancreas. The second dataset is the Task07 pancreas from the Medical Segmentation Decathlon challenge\footnote{\url{http://medicaldecathlon.com}} (MSD)~\cite{simpson2019large}. This dataset contains 281 portal venous phase CT scans with manual labels for the pancreas and pancreatic tumors (intraductal mucinous neoplasms, pancreatic neuroendocrine tumors, or pancreatic ductal adenocarcinoma). The third dataset is from the MICCAI Multi-Atlas Labeling Beyond the Cranial Vault challenge (Synapse)\footnote{\url{https://www.synapse.org/\#!Synapse:syn3193805/wiki/217785}}~\cite{synapse}. This dataset contains 30 portal venous contrast phase CT scans with manual labels for 13 abdominal organs includes the pancreas. We only keep the pancreas labels for the third dataset and discard the labels for the other 12 organs. We randomly shuffled the three datasets separately and split them into training, validation, and testing sets with the ratio of 60\%, 20\%, and 20\%. Among the total 231 training cases, 165 cases have both pancreas and pancreatic tumor labels. 

\subsection{Experimental details}
We use NVIDIA Clara Train SDK 3.1\footnote{\url{https://docs.nvidia.com/clara/clara-train-archive/3.1/index.html}} as the federated learning framework. During the experimentation, the server and associated clients are physically on the same machine and running in individual  Docker containers. The server has no access to GPU, and each client has one V100 32GB GPU. We run the experiments on two machines, the first one is a DGX-Station with 20 CPU cores, 256GB system memory, and 4 V100 32GB GPUs, and the second one is a DGX-1 with 40 CPU cores, 512GB system memory, and 8 V100 GPUs.
All CT volumes were resampled to isotropic spacing with $1.0 \times 1.0 \times 1.0$ $mm^3$. To ensure the CT volumes were in the same orientation, we arranged the voxel axes as close as possible to RAS+ orientation. The Hounsfield unit (HU) intensity in the range [-200, 250] HU were rescaled and clipped into [-1, 1]. We used a network found by coarse-to-fine architecture search (C2FNAS)~\cite{Yu2019-qi} using a TensorFlow implementation in all experiments. The training loss function is the sum of Dice loss and cross entropy. The Adam optimizer with cosine annealing learning rate scheduler was adopted with the initial learning rate $5\times10^{-4}$. The input patch size of our network is $96 \times96\times96$. The total round number of FL was 60 with local epoch number of 10. The minimum client number was 3. Despite running FL in simulation on public datasets, we employed a percentile sharing protocol as a privacy-preserving measure~\cite{li2019Privacy}.  
We only share 25\% of the model updates with the largest absolute values to ensure that our approach could be employed in a real-world setting.
\subsection{Results}
\begin{figure}[tb]
\centering
\begin{tabular}{cc}
    \includegraphics[width=0.45\columnwidth]{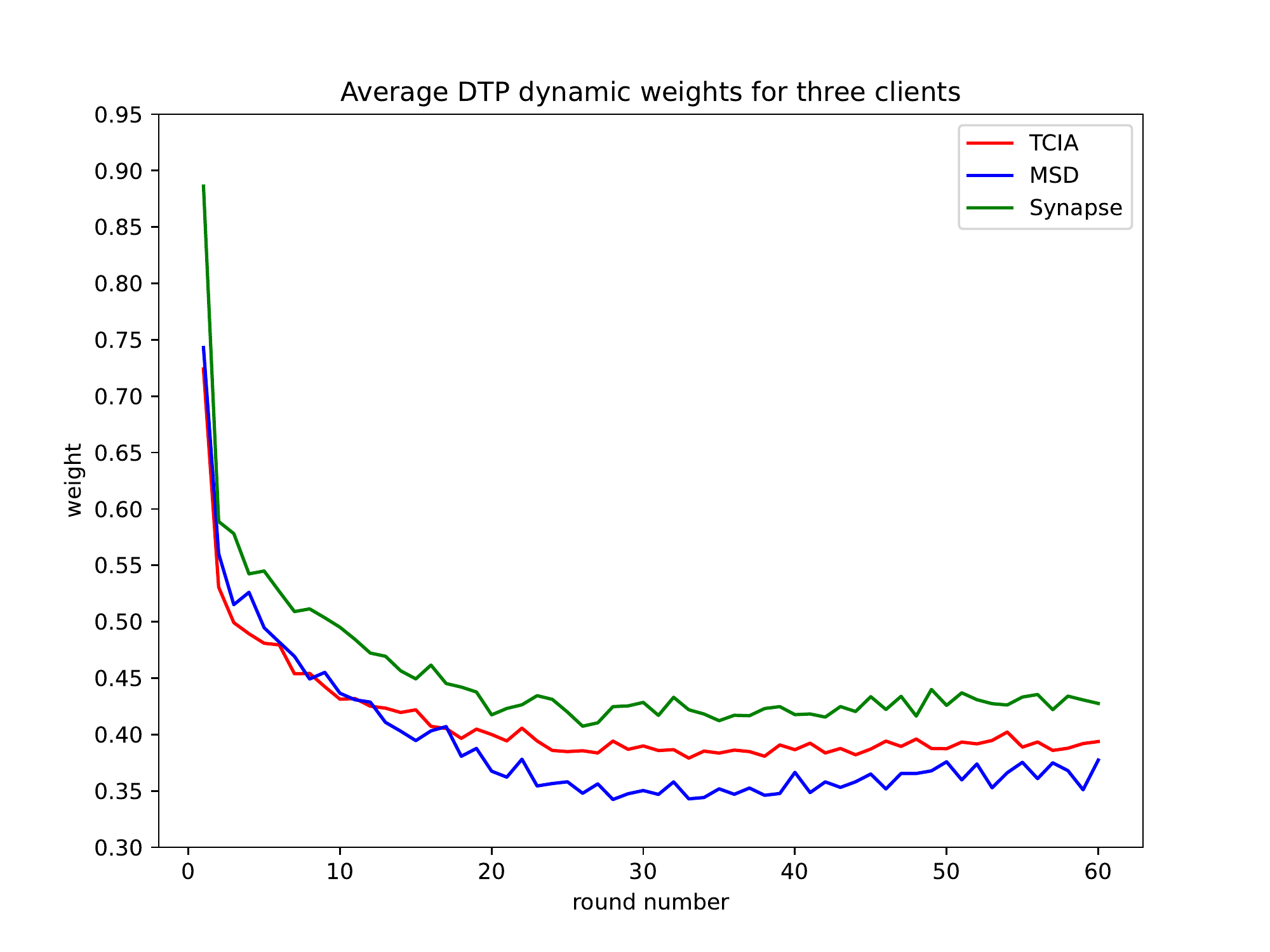} &
    \includegraphics[width=0.45\columnwidth]{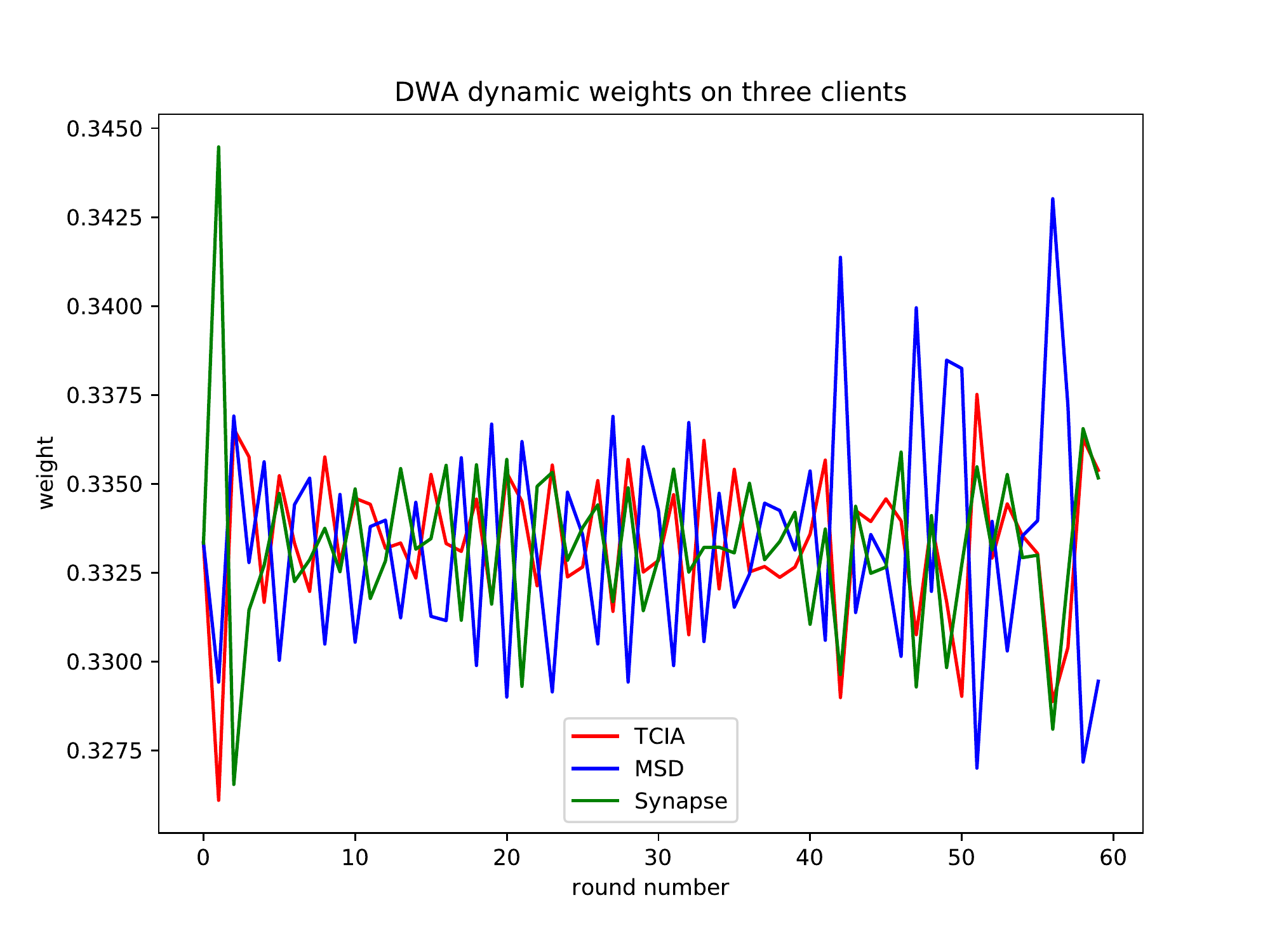} \\
\end{tabular}
\caption{Each client's weight is chosen by DTP and DWA method every round.}
\label{fig:dwa_weights}
\end{figure}
%
Our experimental results include the standalone training model on each dataset (TCIA local, MSD local, and Synapse local) and the FL global best model (determined using the average client validation scores during each FL round) for FedAvg, FedProx, DTP, and DWA. Table \ref{tab:dwa} compares the Dice score across all experiments with different hyperparameter settings. For standalone training model: TCIA local, MSD local, and Synapse local, the performance on other datasets is quite unsatisfactory. FL global models have markedly better generalizability than standalone models. Both DTP and DWA methods rely on precise hyperparameter settings.
For FedAvg, the Dice score on the MSD dataset is highest; however, the performance is not ideal on the Synapse dataset. The average Dice score with FedProx improves over FedAvg. The TCIA dataset and Synapse dataset have the highest Dice score with DWA ($T=2,\xi=2$). The average Dice score with DWA is 4.5\% and 2.9\% higher than FedAvg and FedProx.

\begin{figure}[tbp]
\centering
\begin{tabular}{cc}
\subfloat{\adjincludegraphics[width=0.45\textwidth]{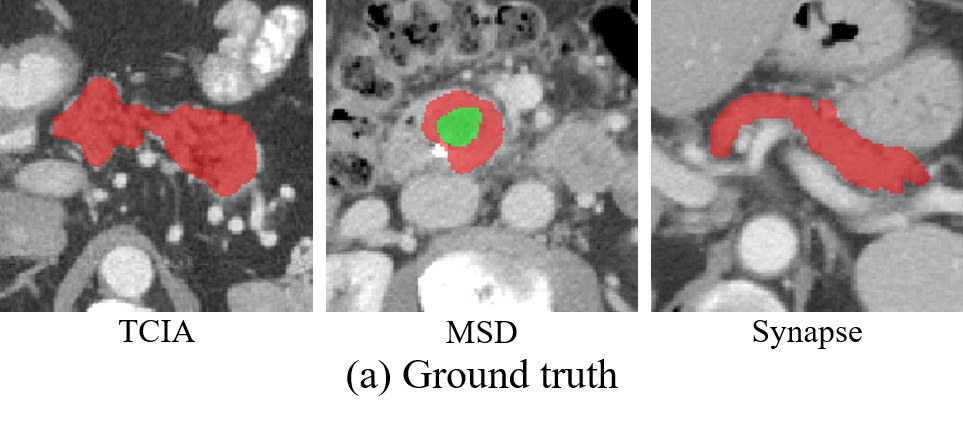}}&
\subfloat{\adjincludegraphics[width=0.45\textwidth]{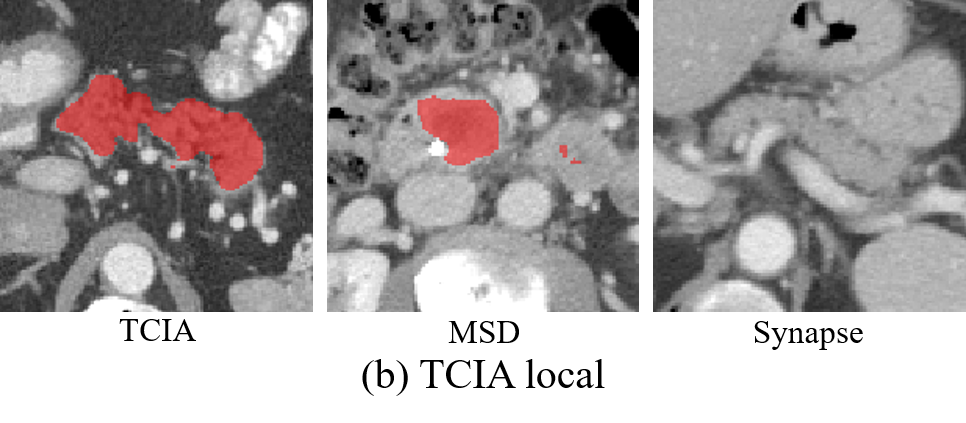}}\\[-1.5ex]

\subfloat{\adjincludegraphics[width=0.45\textwidth]{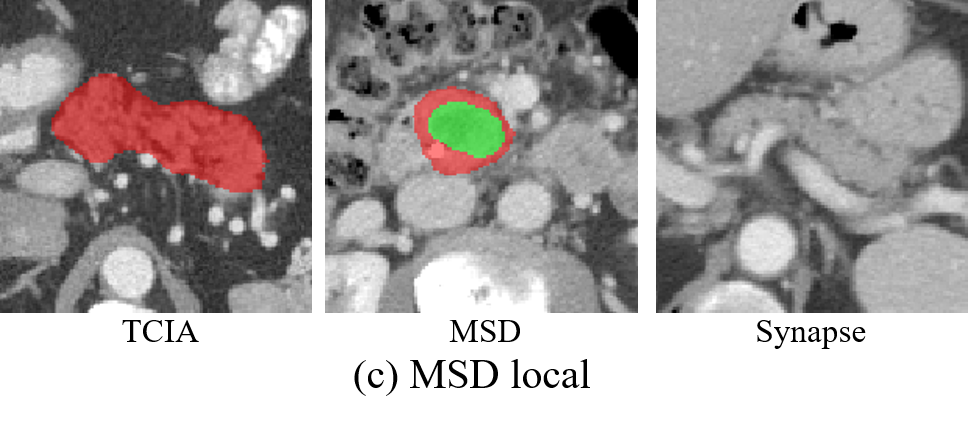}}&
\subfloat{\adjincludegraphics[width=0.45\textwidth]{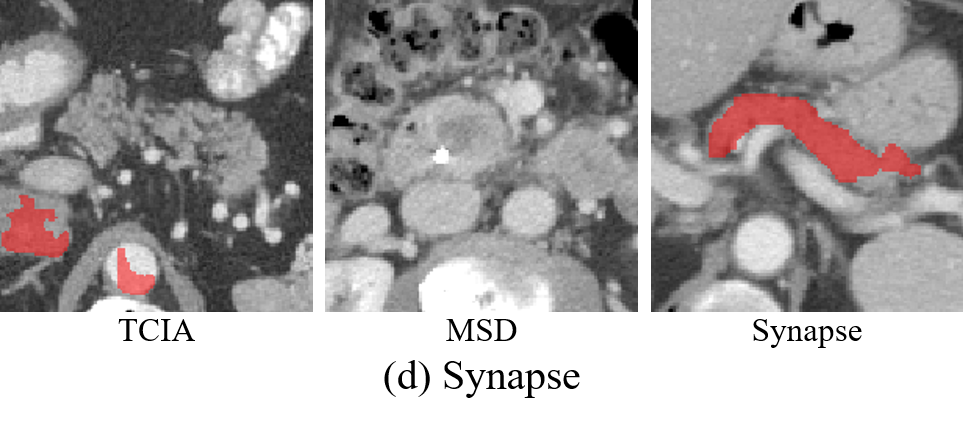}}\\[-1.5ex]
\subfloat{\adjincludegraphics[width=0.45\textwidth]{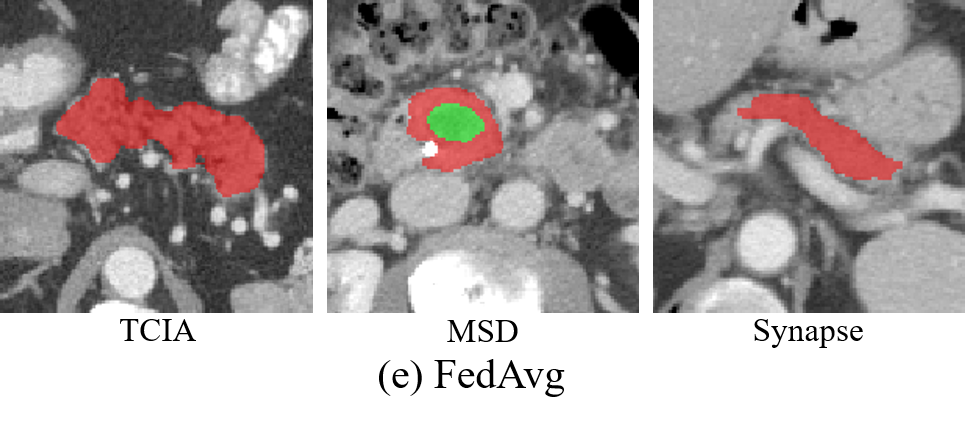}}&
\subfloat{\adjincludegraphics[width=0.45\textwidth]{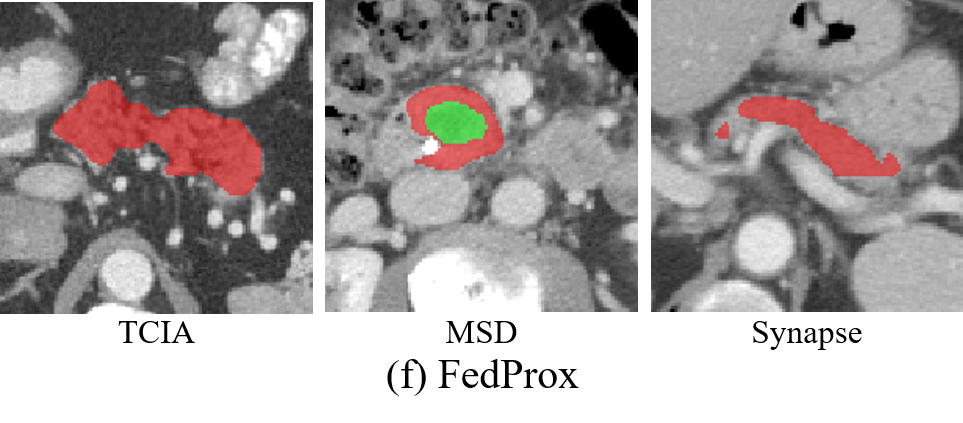}}\\[-1.5ex]
\subfloat{\adjincludegraphics[width=0.45\textwidth]{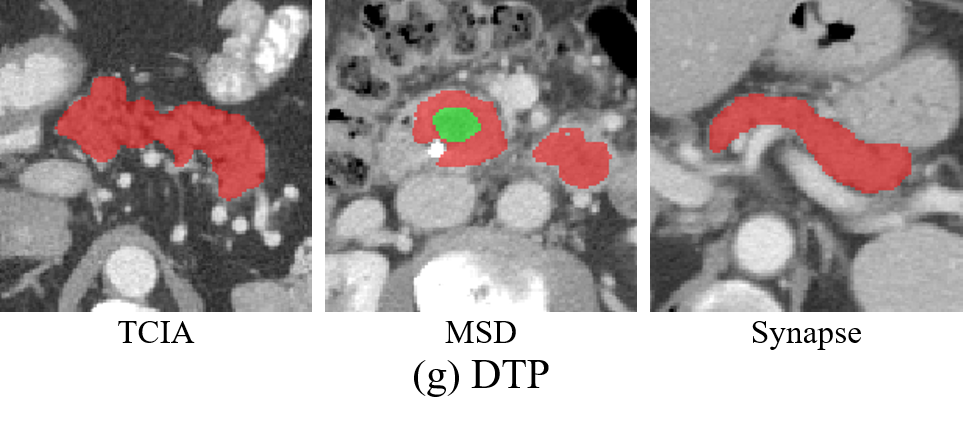}}&
\subfloat{\adjincludegraphics[width=0.45\textwidth]{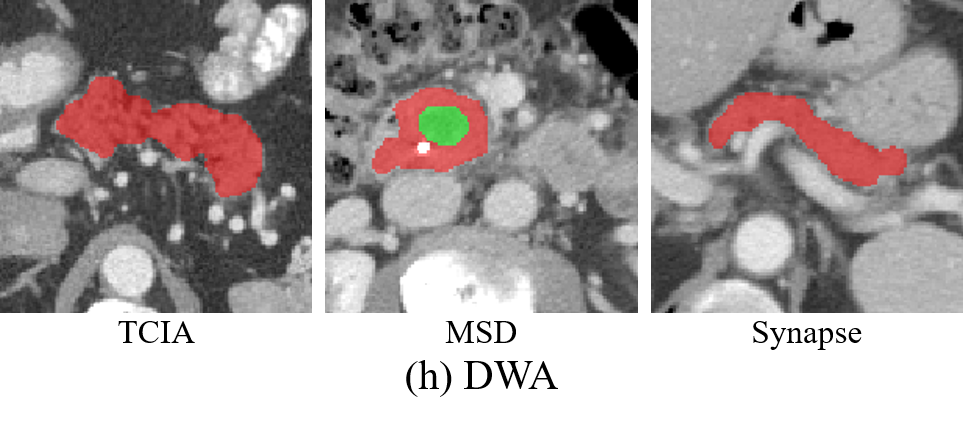}}\\
\end{tabular}
\caption{Examples of pancreas and tumor segmentation on (b) TCIA local, (c) MSD local, (d) Synapse local, and on FL server global models with (e) FedAvg \cite{McMahan2017}, (f) FedProx \cite{li2018federated}, (g) DTP and (h) DWA of TCIA, MSD and Synapse dataset, respectively. }
\label{fig:result}
\end{figure}

Axial visualizations of the segmentation results are shown in Fig.\ref{fig:result}. The performance of using FedAvg and FedProx is close: the segmentation results on TCIA and MSD data are acceptable but not ideal on relatively small Synapse data. In contrast, the DWA method shows a more stable performance on three different datasets. We visualize the dynamically chosen weights by the DTP, and DWA approaches in Fig. \ref{fig:dwa_weights}.
%
\begin{table}[tbp]
\centering
\caption{Comparison of Dice Score for the pancreas and tumor segmentation on \textit{local models} which trained from scratch with single datasets (TCIA, MSD and Synapse); and on \textit{FL server best global model} with FedAvg, FedProx, DTP and DWA optimization. Best scores are shown in \textbf{blod}. \textit{Italic} scores indicate the local models performance on its own test data. Non-italic numbers show the lack of generalizability of local models evaluated on other clients' test data.}
\begin{tabular}{l|cccc|c}
\hline
 & \textbf{TCIA} & \multicolumn{2}{c}{\textbf{MSD}} & \textbf{Synapse} & \textbf{All} \\
 & \textbf{pancreas} & \textbf{pancras} & \textbf{tumor} & \textbf{pancreas} & \textbf{avg} \\ \hline
TCIA local ($N_{train}=48$) & \textit{79.4}\% & 71.8\% & 0.0\% & 5.0\% & 40.1\% \\
MSD local ($N_{train}=165$) & 61.9\% & \textit{77.8}\% & \textit{31.1}\% & 4.4\% & 40.3\% \\
Synapse local ($N_{train}=18$) & 9.8\% & 0.4\% & 0.0\% & \textit{61.1}\% & 23.7\% \\ \hline
FedAvg \cite{McMahan2017} & 80.6\% & \textbf{75.1\%} & \textbf{20.2\%} & 42.6\% & 56.9\% \\
FedProx \cite{li2018federated}& 80.6\% & 75.0\% & 19.5\% & 47.6\% & 58.5\% \\ \hline
DTP($\gamma=1,\alpha=0.9,r=1$)& 64.1\% & 54.8\% & 14.4\% & 34.5\% & 44.4\% \\
DTP($\gamma=2,\alpha=0.9,r=1$) & 46.0\% & 57.3\% & 12.4\% & 27.6\% & 36.1\% \\
DTP($\gamma=1,\alpha=0.5,r=1$) & 64.4\% & 57.2\% & 12.4\% & 39.2\% & 46.1\% \\
DTP($\gamma=1,\alpha=0.5,r=2$) & 64.3\% & 55.7\% & 12.4\% & 39.3\% & 45.9\% \\ \hline
DWA($T=1$) & 65.3\% & 59.8\% & 9.5\% & 49.8\% & 49.9\% \\
DWA($T=1.5$) & 76.1\% & 71.7\% & 13.7\% & 53.0\% & 57.3\% \\
DWA($T=2$) & 78.2\% & 72.4\% & 6.8\% & 56.2\% & 58.0\% \\
DWA($T=2, \xi=2$) & \textbf{80.9\%} & 73.4\% & 13.9\% & \textbf{59.6\%} & \textbf{61.4\%} \\
DWA($T=2, \xi=3$) & 68.7\% & 59.8\% & 7.7\% & 39.3\% & 47.3\% \\ \hline
\end{tabular}
\label{tab:dwa}
\end{table}

\section{Discussion}
As shown in Table \ref{tab:dwa}, the FedAvg is the standard federated learning baseline to compare with other methods. Three local models are standalone-training results for their corresponding datasets. The global model resulting from FedAvg performs well in the TCIA dataset, and the MSD pancreas compares to local models. Moreover, for the MSD tumor and Synapse dataset, although the performance is not as high as corresponding local models, there is still a significant improvement to other local models, indicating the improved generalizability of the global models.
The FedProx model shows similar performance as the FedAvg model. In the MSD dataset, the average Dice score of the pancreas and tumor is slightly lower than the result of the FedAvg model. However, the average Dice score of the Synapse dataset is significantly higher than the result of the FedAvg model.
The performance of DTP models is generally lower than the FedAvg baseline and DWA results. Nonetheless, in most settings, DTP models still outperform the local models. Furthermore, compared to the DWA results, the performance on MSD tumors is more consistent.
%
Both FedAvg and FedProx are commonly used in FL, and our experiments suggest that both methods already provide a strong baseline performance even on heterogeneous datasets.

In DTP, the dynamic prioritization weight focuses on the most challenging tasks by adjusting the magnitude of the loss. However, each client only calculates the prioritization weights using local batch data. The lack of a global perspective of the training can therefore limit the performance of DTP. Also, DTP scales the magnitude of the loss, disrupting the optimization and increasing the need for further hyperparameter tuning.
In contrast, with most DWA configurations, the Synapse dataset's performance is markedly higher than the FedAvg baseline. The results show that DWA can outperform both FedAvg and FedProx on average with properly selected hyperparameters.

\section{Conclusion}
In this work, we investigated two multi-task optimization methods for FL in medical imaging with heterogeneous datasets: DTP and DWA. The application of both methods was inspired by the analogy and similarity between FL and multi-task learning. We evaluated each method within an FL framework and compared the global model performance with FedAvg and FedProx. The Dice of DTP is lower than FedAvg and FedProx, likely because of limited manual tuning. However, the global model from DTP still outperforms the local models. DWA model aggregation method shows significant improvement, especially on the Synapse client whose training data is relatively smaller than other two clients. 

\paragraph{Acknowledgement} Parts of this research was supported by the MEXT/JSPS KAKENHI (894030, 17H00867).
%
%
\bibliographystyle{splncs04}
\bibliography{mybibliography}

\begin{thebibliography}{10}
\providecommand{\url}[1]{\texttt{#1}}
\providecommand{\urlprefix}{URL }
\providecommand{\doi}[1]{https://doi.org/#1}

\bibitem{czeizler2020using}
Czeizler, E., Wiessler, W., Koester, T., Hakala, M., Basiri, S., Jordan, P.,
  Kuusela, E.: Using federated data sources and varian learning portal
  framework to train a neural network model for automatic organ segmentation.
  Physica Medica  \textbf{72},  39--45 (2020)

\bibitem{dou2021Federated}
Dou, Q., So, T.Y., Jiang, M., Liu, Q., Vardhanabhuti, V., Kaissis, G., Li, Z.,
  Si, W., Lee, H.H.C., Yu, K., Feng, Z., Dong, L., Burian, E., Jungmann, F.,
  Braren, R., Makowski, M., Kainz, B., Rueckert, D., Glocker, B., Yu, S.C.H.,
  Heng, P.A.: Federated deep learning for detecting {{COVID}}-19 lung
  abnormalities in {{CT}}: A privacy-preserving multinational validation study.
  npj Digit. Med.  \textbf{4}(1), ~60 (Dec 2021).
  \doi{10.1038/s41746-021-00431-6}

\bibitem{flores2021Federated}
Flores, M., Dayan, I., Roth, H., Zhong, A., Harouni, A., Gentili, A., Abidin,
  A., Liu, A., Costa, A., Wood, B., Tsai, C.S., Wang, C.H., Hsu, C.N., Lee, C.,
  Ruan, C., Xu, D., Wu, D., Huang, E., Kitamura, F., Lacey, G., Corradi,
  G.C.d.A., Shin, H.H., Obinata, H., Ren, H., Crane, J., Tetreault, J., Guan,
  J., Garrett, J., Park, J.G., Dreyer, K., Juluru, K., Kersten, K., Rockenbach,
  M.A.B.C., Linguraru, M., Haider, M., AbdelMaseeh, M., Rieke, N., Damasceno,
  P., e~Silva, P.M.C., Wang, P., Xu, S., Kawano, S., Sriswa, S., Park, S.Y.,
  Grist, T., Buch, V., Jantarabenjakul, W., Wang, W., Tak, W.Y., Li, X., Lin,
  X., Kwon, F., Gilbert, F., Kaggie, J., Li, Q., Quraini, A., Feng, A., Priest,
  A., Turkbey, B., Glicksberg, B., Bizzo, B., Kim, B.S., {Tor-Diez}, C., Lee,
  C.C., Hsu, C.J., Lin, C., Lai, C.L., Hess, C., Compas, C., Bhatia, D.,
  Oermann, E., Leibovitz, E., Sasaki, H., Mori, H., Yang, I., Sohn, J.H.,
  Murthy, K.N.K., Fu, L.C., de~Mendon{\c c}a, M.R.F., Fralick, M., Kang, M.K.,
  Adil, M., Gangai, N., Vateekul, P., Elnajjar, P., Hickman, S., Majumdar, S.,
  McLeod, S., Reed, S., Graf, S., Harmon, S., Kodama, T., Puthanakit, T.,
  Mazzulli, T., Lavor, V.d.L., Rakvongthai, Y., Lee, Y.R., Wen, Y.: Federated
  {{Learning}} used for predicting outcomes in {{SARS}}-{{COV}}-2 patients
  (2021). \doi{10.21203/rs.3.rs-126892/v1}

\bibitem{guo2018dynamic}
Guo, M., Haque, A., Huang, D.A., Yeung, S., Fei-Fei, L.: Dynamic task
  prioritization for multitask learning. In: Proceedings of the European
  Conference on Computer Vision (ECCV). pp. 270--287 (2018)

\bibitem{synapse}
Landmanm, B., Xu, Z., Igelsias, Juan, E., Styner, M., Langerak, Thomas, R.,
  Klein, A.: 2015 {MICCAI} multi-atlas labeling beyond the cranial vault –
  workshop and challenge  (2015). \doi{10.7303/syn3193805}

\bibitem{li2018federated}
Li, T., Sahu, A.K., Zaheer, M., Sanjabi, M., Talwalkar, A., Smith, V.:
  Federated optimization in heterogeneous networks. arXiv preprint
  arXiv:1812.06127  (2018)

\bibitem{li2019Privacy}
Li, W., Milletar{\`i}, F., Xu, D., Rieke, N., Hancox, J., Zhu, W., Baust, M.,
  Cheng, Y., Ourselin, S., Cardoso, M.J., Feng, A.: Privacy-preserving
  federated brain tumour segmentation. In: Suk, H.I., Liu, M., Yan, P., Lian,
  C. (eds.) Machine Learning in Medical Imaging. pp. 133--141. {Springer
  International Publishing}, {Cham} (2019)

\bibitem{Liu_2019_CVPR}
Liu, S., Johns, E., Davison, A.J.: End-to-end multi-task learning with
  attention. In: Proceedings of the IEEE/CVF Conference on Computer Vision and
  Pattern Recognition (CVPR) (2019)

\bibitem{man2019deep}
Man, Y., Huang, Y., Feng, J., Li, X., Wu, F.: Deep q learning driven ct
  pancreas segmentation with geometry-aware u-net. IEEE transactions on medical
  imaging  \textbf{38}(8),  1971--1980 (2019)

\bibitem{McMahan2017}
McMahan, H.B., Moore, E., Ramage, D., Hampson, S., y~Arcas, B.A.:
  Communication-efficient learning of deep networks from decentralized data.
  In: AISTATS (2017)

\bibitem{oktay2018attention}
Oktay, O., Schlemper, J., Folgoc, L.L., Lee, M., Heinrich, M., Misawa, K.,
  Mori, K., McDonagh, S., Hammerla, N.Y., Kainz, B., et~al.: Attention u-net:
  Learning where to look for the pancreas. arXiv preprint arXiv:1804.03999
  (2018)

\bibitem{roth2015deeporgan}
Roth, H.R., Lu, L., Farag, A., Shin, H.C., Liu, J., Turkbey, E.B., Summers,
  R.M.: Deeporgan: Multi-level deep convolutional networks for automated
  pancreas segmentation. In: International conference on medical image
  computing and computer-assisted intervention. pp. 556--564. Springer (2015)

\bibitem{sheller2019Multiinstitutional}
Sheller, M.J., Reina, G.A., Edwards, B., Martin, J., Bakas, S.:
  Multi-institutional {{Deep Learning Modeling Without Sharing Patient Data}}:
  {{A Feasibility Study}} on {{Brain Tumor Segmentation}}. In: Brainlesion:
  {{Glioma}}, {{Multiple Sclerosis}}, {{Stroke}} and {{Traumatic Brain
  Injuries}}. vol. 11383, pp. 92--104. {Springer International Publishing},
  {Cham} (2019). \doi{10.1007/978-3-030-11723-8\_9}

\bibitem{simpson2019large}
Simpson, A.L., Antonelli, M., Bakas, S., Bilello, M., Farahani, K.,
  Van~Ginneken, B., Kopp-Schneider, A., Landman, B.A., Litjens, G., Menze, B.,
  et~al.: A large annotated medical image dataset for the development and
  evaluation of segmentation algorithms. arXiv preprint arXiv:1902.09063
  (2019)

\bibitem{smith2017federated}
Smith, V., Chiang, C.K., Sanjabi, M., Talwalkar, A.: Federated multi-task
  learning. In: Proceedings of the 31st International Conference on Neural
  Information Processing Systems. pp. 4427--4437 (2017)

\bibitem{wang2020Automated}
Wang, P., Shen, C., Roth, H.R., Yang, D., Xu, D., Oda, M., Misawa, K., Chen,
  P.T., Liu, K.L., Liao, W.C., Wang, W., Mori, K.: Automated pancreas
  segmentation using multi-institutional collaborative deep learning. In:
  Domain Adaptation and Representation Transfer, and Distributed and
  Collaborative Learning. pp. 192--200. Springer International Publishing, Cham
  (2020)

\bibitem{xia2021AutoFedAvg}
Xia, Y., Yang, D., Li, W., Myronenko, A., Xu, D., Obinata, H., Mori, H., An,
  P., Harmon, S., Turkbey, E., Turkbey, B., Wood, B., Patella, F., Stellato,
  E., Carrafiello, G., Ierardi, A., Yuille, A., Roth, H.: Auto-{{FedAvg}}:
  {{Learnable Federated Averaging}} for {{Multi}}-{{Institutional Medical Image
  Segmentation}}  (2021)

\bibitem{yang2021federated}
Yang, D., Xu, Z., Li, W., Myronenko, A., Roth, H.R., Harmon, S., Xu, S.,
  Turkbey, B., Turkbey, E., Wang, X., et~al.: Federated semi-supervised
  learning for covid region segmentation in chest ct using multi-national data
  from china, italy, japan. Medical image analysis  \textbf{70},  101992 (2021)

\bibitem{Yu2019-qi}
Yu, Q., Yang, D., Roth, H., Bai, Y., Zhang, Y., Yuille, A.L., Xu, D.: {C2FNAS}:
  {Coarse-to-Fine} neural architecture search for {3D} medical image
  segmentation  (Dec 2019)

\bibitem{zhou2017fixed}
Zhou, Y., Xie, L., Shen, W., Wang, Y., Fishman, E.K., Yuille, A.L.: A
  fixed-point model for pancreas segmentation in abdominal ct scans. In:
  International conference on medical image computing and computer-assisted
  intervention. pp. 693--701. Springer (2017)

\end{thebibliography}

\end{document}